# Enhanced Bilingual Evaluation Understudy

Krzysztof Wołk, Krzysztof Marasek
Department of Multimedia
Polish Japanese Institute of Information Technology, Warsaw, POLAND
kwolk@pjwstk.edu.pl

**Abstract** - Our research extends the Bilingual Evaluation Understudy (BLEU) evaluation technique for statistical machine translation to make it more adjustable and robust. We intend to adapt it to resemble human evaluation more. We perform experiments to evaluate the performance of our technique against the primary existing evaluation methods. We describe and show the improvements it makes over existing methods as well as correlation to them. When human translators translate a text, they often use synonyms, different word orders or style, and other similar variations. We propose an SMT evaluation technique that enhances the BLEU metric to consider variations such as those.

## I. INTRODUCTION

To make progress in Statistical Machine Translation (SMT), the quality of its results must be evaluated. It has been recognized for quite some time that using humans to evaluate SMT approaches is very expensive and time-consuming. [1] As a result, human evaluation cannot keep up with the growing and continual need for SMT evaluation. This led to the recognition that the development of automated SMT evaluation techniques is critical. [1, 2]

Evaluation is particularly crucial for translation between diverse language pairs, such as Polish and English. Polish has complex declension, 7 cases, 15 gender forms, and complicated grammatical construction procedures. This leads to a very large Polish vocabulary and great complexity in data requirements for SMT. Meanwhile, the order of subjects, verbs, and objects is not important to determine the meaning of a Polish sentence. Instead, many variations of word order mean the same thing in this language.

Unlike Polish, the English language does not have declensions. In addition, word order, esp. the Subject-Verb-Object (SVO) pattern, is absolutely crucial to determining the meaning of an English sentence.

These differences in the Polish and English languages lead to great translation complexity. In addition, the lack of lexical data availability and phrase models only further complicates SMT between those languages.

In [2] Reeder compiled an initial list of SMT evaluation metrics. Further research has led to the development of newer metrics. Prominent metrics include: Bilingual Evaluation Understudy (BLEU); the National Institute of Standards and Technology (NIST) metric; Translation Error Rate (TER), the Metric for Evaluation of Translation with Explicit Ordering (METEOR); Length Penalty, Precision, n-gram Position difference Penalty and Recall (LEPOR); and the Rank-based Intuitive Bilingual Evaluation Score (RIBES).

This paper presents extensions to existing SMT evaluation metrics. Section 2 describes the existing evaluation techniques. Our enhanced method is discussed in Section 3. Section 4 describes the experiments we performed to compare our method with existing evaluation methods. Section 5 discusses the results and future potential research in this area.

## II. EXISTING EVALUATION TECHNIQUES

This section describes existing SMT evaluation techniques.

### A. BLEU Metric

BLEU was developed based on a premise similar to that used for speech recognition, described in [3] as: "The closer a machine translation is to a professional human translation, the better it is." So, the BLEU metric is designed to measure how close SMT output is to that of human reference translations. It is important to note that translations, SMT or human, may differ significantly in word usage, word order, and phrase length. [3]

To address these complexities, BLEU attempts to match variable length phrases between SMT output and reference translations. Weighted match averages are used to determine the translation score. [4]

A number of variations of the BLEU metric exist. However, the basic metric requires calculation of a brevity penalty $P_B$, which is calculated as follows:

$$P_B = \begin{cases} 1, & c > r \\ e^{(1-r/c)}, & c \leq r \end{cases} \quad (1)$$

where $r$ is the length of the reference corpus, and candidate (reference) translation length is given by $c$. [4]

The basic BLEU metric is then determined as shown in [4]:

$$BLEU = P_B \exp(\sum_{n=0}^{N} w_n \log p_n) \quad (2)$$

where $w_n$ are positive weights summing to one, and the $n$-gram precision $p_n$ is calculated using $n$-grams with a maximum length of N.

There are several other important features of BLEU. First, word and phrase position within text are not evaluated by this metric. To prevent SMT systems from artificially inflating their scores by overuse of words known with high confidence, each candidate word is constrained by the word count of the corresponding reference translation. A geometric mean of individual sentence scores, with consideration of the brevity penalty, is then calculated for the entire corpus. [4]

## B. NIST Metric

The NIST metric was designed to improve BLEU by rewarding the translation of infrequently used words. This was intended to further prevent inflation of SMT evaluation scores by focusing on common words and high confidence translations. As a result, the NIST metric uses heavier weights for rarer words. The final NIST score is calculated using the arithmetic mean of the *n*-gram matches between SMT and reference translations. In addition, a smaller brevity penalty is used for smaller variations in phrase lengths. The reliability and quality of the NIST metric has been shown to be superior to the BLEU metric. [5]

## C. Translation Edit Rate (TER)

Translation Edit Rate (TER) was designed to provide a very intuitive SMT evaluation metric, requiring less data than other techniques while avoiding the labor intensity of human evaluation. It calculates the number of edits required to make a machine translation match exactly to the closest reference translation in fluency and semantics. [6, 7]

Calculation of the TER metric is defined in [6]:

$$TER = \frac{E}{w_R} \qquad (3)$$

where $E$ represents the minimum number of edits required for an exact match, and the average length of the reference text is given by $w_R$. Edits may include the deletion of words, word insertion, word substitutions, as well as changes in word or phrase order. [6]

## D. METEOR Metric

The Metric for Evaluation of Translation with Explicit Ordering (METEOR) is intended to take several factors that are indirect in BLEU into account more directly. Recall (the proportion of matched n-grams to total reference n-grams) is used directly in this metric. In addition, METEOR explicitly measures higher order n-grams, considers word-to-word matches, and applies arithmetic averaging for a final score. Best matches against multiple reference translations are used.[8]

The METEOR method uses a sophisticated and incremental word alignment method that starts by considering exact word-to-word matches, word stem matches, and synonym matches. Alternative word order similarities are then evaluated based on those matches.

Calculation of precision is similar in the METEOR and NIST metrics. Recall is calculated at the word level. To combine the precision and recall scores, METEOR uses a harmonic mean. METEOR rewards longer *n*-gram matches. [8]

The METEOR metric is calculated as shown in [8]:

$$METEOR = \left(\frac{10\,P\,R}{R+9\,P}\right)(1 - P_M) \qquad (4)$$

where the unigram recall and precision are given by $R$ and $P$, respectively. The brevity penalty $P_M$ is determined by:

$$P_M = 0.5 \left(\frac{C}{M_U}\right) \qquad (5)$$

where $M_U$ is the number of matching unigrams, and $C$ is the minimum number of phrases required to match unigrams in the SMT output with those found in the reference translations.

## E. LEPOR

Some SMT evaluation metrics perform well on certain languages but poorly on others. The LEPOR metric was specifically designed to address this problem of language bias. As a result, it does not rely on linguistic features or data specific to a particular language. [9]

This metric increases the penalty for translations shorter or longer than the reference translations. LEPOR also institutes an n-gram word order penalty, and combines the penalties with precision and recall measures. [9, 10]

The basic LEPOR metric is calculated by [9]:

LEPOR = LP x NPosPenal x Harmonic(αR, βP)

where *LP* is the length penalty, *NPosPen* is the n-gram position difference penalty, *R* is recall, *P* is precision, and α and β are adjustable weights.

The length penalty is defined by [10]:

$$LP = \begin{cases} e^{1-\frac{r}{c}} & if\ c < r \\ 1 & if\ c = r \\ e^{1-\frac{c}{r}} & if\ c > r \end{cases} \qquad (6)$$

where $c$ is the average length of SMT sentences and $r$ is the average length of reference translation sentences.

The normalized n-gram penalty is calculated by:

$$NPosPenal = e^{-NPD} \qquad (7)$$

where *NPD* is n-gram position difference penalty. Details of the calculation of *NPD* may be found in [9].

## F. RIBES

The focus of the RIBES metric is word order. It uses rank correlation coefficients based on word order to compare SMT and reference translations. The primary rank correlation coefficients used are Spearman's ρ, which measures the distance of differences in rank, and Kendall's τ, which measures the direction of differences in rank. [11]

These rank measures can be normalized to ensure positive values [11]:

Normalized Spearman's ρ (NSR) = (ρ + 1)/2
Normalized Kendall's τ (NKT) = (τ + 1)/2

These measures can be combined with precision *P* and modified to avoid overestimating the correlation of only corresponding words in the SMT and reference translations:

NSR $P^α$ and NKT $P^α$

where α is a parameter in the range $0 \leq α \leq 1$.

## III. ENHANCED EVALUATION TECHNIQUE

When human translators translate a text, they often use synonyms, different word orders or style, and other similar variations. We propose an SMT evaluation technique that enhances the BLEU metric to consider variations such as those. First, we will review some key features of the BLEU metric. Then, we will describe our technique.

*A. Key Features of BLEU*

In the BLEU metric, scores are calculated for individual translated segments (generally sentences). Those scores are then averaged over the entire corpus to reach an estimate of the translation's overall quality. The BLEU score is always a number between 0 and 1.

BLEU uses a modified form of precision to compare a candidate translation against multiple reference translations. An over-simplified example of this is:
  Test Phrase: "the the the the the the the"
  Reference 1 Phrase: "the cat is on the mat"
  Reference 2 Phrase: "there is a cat on the mat"

In this example, precision score is the number of words in the test phrase that are found in the reference phrases (7) divided by the total number of words in the test phrase. This would yield a perfect precision score of 1.

This is a perfect score for a poor translation. BLEU solves this problem with a simple modification: for each word in a test phrase, it uses the minimum of the test phrase word count and the reference word count.

If we have more than one reference, BLEU first takes the maximum word count of all references and compares it with the test phrase word count. For the example above:
  Count("the" in Test) = 7
  Count("the" in Ref1) = 2
  Count("the" in Ref2) = 1

BLEU first determines 2 as the maximum matching word count among all references. It then chooses the minimum of that value and the test phrase word count:
  min(7, 2) = 2

BLEU calculates this minimum for each non-repeated word in the test phrase. In our example, it calculates this minimum value just one time for word "the". The final score is determined by the sum of the minimum values for each word divided by the total number of words in the test phrase:
$$Final\ Score = \left(\frac{2}{7}\right) = 0.2857 \qquad (8)$$

Another problem with BLEU scoring is that it tends to favor translations of short phrases, due to dividing by the total number of words in the test phrase.

For example, consider this translation for above example:
  Test Phrase: "the cat" : score = (1+1)/2 = 1
  Tes Phrase: "the" : score = 1/1 = 1

BLEU uses a brevity penalty, as previously described, to prevent very short translations. BLEU also uses n-grams. For example, for this test phrase: "the cat is here" with n-grams, we have:
  1-gram: "the", "cat", "is", "here"
  2-gram: "the cat", "cat is", "is here"
  3-gram: "the cat is", "cat is here"
  4-gram: "the cat is here"

For the reference phrase "the cat is on the mat", we have, for example, the following 2-grams: "the cat", "cat is", "is on", "on the", "the mat".

BLEU calculates the score for each of the n-grams. So in calculation of the following 2-grams:
  Test 2-grams: "the cat", "cat is", "is here"
  Reference 2-grams: "the cat", "cat is", "is on", "on the", "the mat"
it takes:
  "the cat": 1
  "cat is": 1
  "is here": 0
  2-grams score = (1+1+0)/3 = 2/3

*B. Enhanced Metric*

We now discuss enhancements to the BLEU metric. In particular, our enhanced metric rewards synonyms and rare word translations, while modifying the calculation of cumulative scores.

*a. Consideration of Synonyms*

In our enhanced metric, we would like to reward matches of synonyms, since the correct meaning is still conveyed.

Consider this test phrase: "this is a exam" and this reference phrase: "this is a quiz"

The BLEU score is calculated as follows:
  BLEU = (1+1+1+0)/4 = 3/4 = 0.75

BLEU does not count the word "exam" as a match, because it does not find it in the reference phrase. However, this word is not a bad choice. In our method, we want to score the synonym "exam" higher than zero and lower than the exact word "quiz".

To do this, for each word in a test phrase we try to find its synonyms. We check for an exact word match and for all test phrase synonyms to find the closest words to the reference.

For example, for the phrases:
  Test: "this is a exam"
  Reference: "this is a quiz"

"exam" has some synonyms, e.g., "test", "quiz", and "examination."

We check each synonym in the reference. If a synonym has a greater number of matches in the reference, we replace it with the original word.

In this example we replace "quiz" to reach this test sentence: "this is a quiz". Which modifies our test phrase to be: "this is a quiz".

We apply the default BLEU algorithm to the modified test phrase and reference phrase, with one difference. The default BLEU algorithm scores this new test phrase as 1.0, but we know that the original test phrase is "this is a exam". So, we would like to give a score higher than 0.75 but less than 1.0 to the test phrase.

During the BLEU evaluation, we check each word for an exact match. If the word is a synonym and not an exact match, we do not give a full score to that word. The score for a synonym will be the default BLEU score for an original word multiplied by a constant (synonym-score).

For example, if this constant equals 0.90, the new score with synonyms is:

$$(1+1+1+0.9)/4 = 3.9/4 = 0.975$$

With this algorithm, we have synonym scores for all n-grams, because in 2-gram we have "a quiz" and in 3-gram,"is a quiz" in both test and reference phrases.

*b. Consideration of Rare Words*

Our algorithm gives extra points to rare word matches. First, it obtains the rare words found in the reference corpus. If we sort all distinct words of the reference with their repetition order (descending), the last words in this list are rare words. The algorithm takes a specific percentage of the whole sorted list as the rare words (rare-words-percent).

When the default BLEU algorithm tries to score a word, if this word is in the rare word list, the score is multiplied by a constant (rare-words-score). This action applies to all n-grams. So, if we have a rare word in a 2-gram, the algorithm increases the score for this 2-gram. For example, if the word "roman" is rare, the "roman empire" 2-gram gets an increased score.

The algorithm is careful that score of each sentence falls within the range of 0.0 and 1.0.

*c. Determination of Cumulative Score*

The cumulative score of our algorithm combines default BLEU scores using logarithms and exponentials as follows:
1. Initialize s = 0
2. For each *i*th-gram:
    a.  s = s + log($B_i$)
    b.  $C_i$ = exp(s / i)

where $B_i$ is the default BLEU score and $C_i$ is the cumulative score.

In addition, we know that:

$$\exp(\log(a) + \log(b)) = a * b$$

and:

$$\exp(\log(a) / b) = a \hat{} (1/b)$$

This simplifies the calculation.

For example, for *i* = 1 to 4:

$C_1 = B_1$
$C_2 = (B_1 * B_2) \hat{} (1/2)$
$C_3 = (B_1 * B_2 * B_3) \hat{} (1/3)$
$C_4 = (B_1 * B_2 * B_3 * B_3) \hat{} (1/4)$

If we have:
$B_1 = 0.70$
$B_2 = 0.55$
$B_3 = 0.37$
$B_4 = 0.28$
then:
$C_1 = 0.70$
$C_2 = 0.62$
$C_3 = 0.52$
$C_4 = 0.44$

The length score (brevity penalty) in our algorithm is calculated as:

len_score = min(0.0, 1 – ref_length / test_ngrams)

and cumulatively:

exp(score / i + len_score)

## IV. EXPERIMENTAL DATA

We conducted experiments to compare performance of our enhanced SMT evaluation metric with that of the most popular metrics: BLEU, NIST, TER, and METEOR for SMT between Polish and English.

The data set used for the experiments was the European Medicines Agency (EMEA) parallel corpus [12].

Table 1 shows the results of our Polish to English translation experiments. Table 2 shows the results of our English to Polish translation experiments. EBLEU column is evaluation with our new metric.

TABLE I. POLISH TO ENGLISH TRANSLATIONS RESULTS

| EXP NO | EBLEU | BLEU | NIST | TER | MET | RIBES |
|---|---|---|---|---|---|---|
| 00 | 70.42 | 70.15 | 10.53 | 29.38 | 82.19 | 83,12 |
| 01 | 63.75 | 64.58 | 9.77 | 35.62 | 76.04 | 72,23 |
| 02 | 70.85 | 71.04 | 10.61 | 28.33 | 82.54 | 82,88 |
| 03 | 70.88 | 71.22 | 10.58 | 28.51 | 82.39 | 83,47 |
| 04 | 76.22 | 76.24 | 10.99 | 24.77 | 85.17 | 85,12 |
| 05 | 70.94 | 71.43 | 10.60 | 28.73 | 82.89 | 83,19 |
| 06 | 73.10 | 71.91 | 10.76 | 26.60 | 83.63 | 84,64 |
| 07 | 70.47 | 71.12 | 10.37 | 29.95 | 84.55 | 76,29 |
| 08 | 71.78 | 71.32 | 10.70 | 27.68 | 83.31 | 83,72 |
| 09 | 70.65 | 71.35 | 10.40 | 29.74 | 81.52 | 77,12 |
| 10 | 71.42 | 70.34 | 10.64 | 28.22 | 82.65 | 83,39 |
| 11 | 73.11 | 72.51 | 10.70 | 28.19 | 82.81 | 80,08 |

TABLE II. ENGLISH TO POLISH TRANSLATIONS RESULTS

| EXP NO | EBLEU | BLEU | NIST | TER | MET | RIBES |
|---|---|---|---|---|---|---|
| 00 | 66.81 | 69.18 | 10.14 | 30.90 | 79.21 | 82,92 |
| 01 | 58.28 | 61.15 | 9.19 | 39.45 | 71.91 | 71,39 |
| 02 | 67.24 | 69.41 | 10.14 | 30.90 | 78.98 | 82,44 |
| 03 | 66.33 | 68.45 | 10.06 | 31.62 | 78.63 | 82,70 |
| 04 | 72.00 | 73.32 | 10.48 | 27.05 | 81.72 | 84,59 |
| 05 | 67.31 | 69.27 | 10.16 | 30.80 | 79.30 | 82,99 |
| 06 | 66.64 | 68.43 | 10.07 | 31.27 | 78.95 | 83,26 |
| 07 | 66.41 | 67.61 | 9.87 | 33.05 | 77.82 | 77,77 |
| 08 | 66.64 | 68.98 | 10.11 | 31.13 | 78.90 | 82,38 |
| 09 | 67.30 | 68.67 | 10.02 | 31.92 | 78.55 | 79,10 |
| 10 | 66.76 | 69.01 | 10.14 | 30.84 | 79.13 | 82,93 |
| 11 | 66.66 | 67.47 | 9.89 | 33.32 | 77.65 | 75,19 |

To better assess the association among the metrics, we use correlation. Correlation measures the association among two or more quantitative or qualitative independent variables. [13] So, we use correlation here to estimate the association between metrics.

The correlation between two arrays of variables X and Y can be calculated using the following formula:

$$Correl(X,Y) = \frac{\sum(x-\bar{x})(y-\bar{y})}{\sqrt{\sum(x-\bar{x})^2 \sum(y-\bar{y})^2}} \quad (9)$$

The correlation output table for the metrics is:

TABLE III. CORRELATION FOR POLISH TO ENGLISH

|  | EBLEU | BLEU | NIST | TER | METEOR | RIBES |
|---|---|---|---|---|---|---|
| EBLEU | 1 | | | | | |
| BLEU | 0.9732 | 1 | | | | |
| NIST | 0.9675 | 0.9158 | 1 | | | |
| TER | -0.9746 | -0.9327 | -0.9909 | 1 | | |
| METEOR | 0.8981 | 0.8943 | 0.8746 | -0.8963 | 1 | |
| RIBES | 0.7570 | 0.6738 | 0.8887 | -0.8664 | 0.6849 | 1 |

Table 3 shows that the NIST metric is in a stronger correlation with EBLEU than with BLEU. Our metric shows more negative association with TER than does BLEU. Our metric shows a stronger correlation with METEOR than does BLEU.

Figure 1 shows the data trends, as well as the association of different variables.

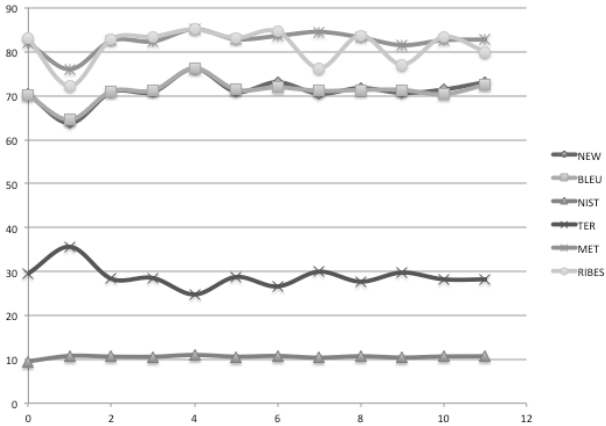

Figure 1. Association of Metric Values.

To confirm the results, we wanted to determine if the same correlation would occur in translations from English to Polish. We took the results and developed an aggregation table in which we merged both tables with results. The aggregation is shown in Table 4.

TABLE IV. AGGREGATION FOR ENGLISH - POLISH

|  | EBLEU | BLEU | NIST | TER | METEOR | RIBES |
|---|---|---|---|---|---|---|
| EBLEU | 1 | | | | | |
| BLEU | 0.9657 | 1 | | | | |
| NIST | 0.9762 | 0.9361 | 1 | | | |
| TER | -0.9666 | -0.9725 | -0.9723 | 1 | | |
| METEOR | 0.9615 | 0.9276 | 0.9653 | -0.9411 | 1 | |
| RIBES | 0.8105 | 0.6989 | 0.9809 | -0.9097 | 0.6849 | 1 |

This shows a stronger correlation between NIST and RIBES and our metric than between NIST or RIBES and BLEU. Our metric has a more negative correlation with TER than does BLEU. Lastly, our metric has a stronger correlation with METEOR than does BLEU.

Finally we wanted to confirm how statistically relevant were the obtained results. To check the correlation coefficiency we additionally counted asymmetric lambda measure of association $\lambda(C|R)$, which is interpreted as the probable improvement in prediction of the column variable Y given knowledge of the row variable X (values given in table). Asymmetric lambda has the range

$$0 \leq \lambda(C|R) \leq 1.$$

It is computed as

$$\lambda(C|R) = \frac{\sum_i r_i - r}{n - r} \quad (10)$$

With

$$var = \frac{n - \sum_i r_i}{(n-r)^3}(\sum_i r_i + r - 2\sum_i (r_i | l_i = l)) \quad (11)$$

Where:

$$r_i = \max_j (n_{ij}) \quad (12)$$

$$r = \max_j (n_{.j}) \quad (13)$$

For this purpose we used IBM's SPPS tool [14]. In our experiments we count EBLEU result as dependent variable (EBLEU is a function of the metrics variable) to every other metric. Using the interpretive guide for measures of association (0.0 = no relationship, ±0.0 to ±0.2 = very weak, ±0.2 to ±0.4 = weak, ±0.4 to ±0.6 = moderate, ±0.6 to ±0.8 = strong, ±0.8 to ±1.0 = very strong, ±1.0 = perfect relationship), our lambda results would be characterized as a very strong relationship if bigger than 0.8 value. [15] Table 5 represents the strength of association between EBLEU and other metrics.

TABLE V. ASSOCIATION STRENGTH

|  | BLEU* | NIST* | TER* | MET* | RIB** |
|---|---|---|---|---|---|
| Symmetric | 0.973 | 0.918 | 0.957 | 0.975 | 0.978 |
| EBLEU Dependent | 0.988 | 0.870 | 0.957 | 01.000 | 1.000 |
| * Dependent | 0.958 | 0.957 | 0.958 | 0.958 | 0.957 |

The lambda results confirm that correlation is very strong for each metric, what is more in the case of the METEOR it is even a perfect relationship.

Lastly we conducted Spearman Correlation [16].

In statistics, its rank is often denoted by the Greek letter $\rho$ (rho) or as $r_s$. It is a nonparametric measure of statistical dependence between two variables. It assesses how well the relationship between two variables can be described using a monotonic function. If there are no repeated data values, a perfect Spearman correlation of +1 or −1 occurs when each of the variables is a perfect monotone function of the other.

Pearson correlation is unduly influenced by outliers, unequal variances, non-normality, and nonlinearity. This latter correlation is calculated by applying the Pearson[1] correlation formula to the ranks of the data rather than to the actual data values themselves. In so doing, many of the distortions that plague the Pearson correlation are reduced considerably.

Aditionally the Pearson correlation measures the strength of linear relationship between *X* and *Y*. In the case of nonlinear, but monotonic relationships, a useful measure is *Spearman's* rank correlation coefficient, *Rho,* which is a *Pearson's* type correlation coefficient computed on the ranks of *X* and *Y* values. It is computed by the following formula (Non-parametric Measures of Bivariate Relationships):

$$Rho = \frac{[1 - 6 \sum (d_i)^2]}{[n(n^2 - 1)]} \quad (14)$$

where

$d_i$ is the difference between the ranks of $X_i$ and $Y_i$.

$r_s = +1$, if there is a perfect agreement between the two sets of ranks.

$r_s = -1$, if there is a complete disagreement between the two sets of ranks.

Spearman's coefficient, like any correlation calculation, is appropriate for both continuous and discrete variables, including ordinal variables. The following Table 6 shows two-tailed Spearman's correlation for EBLEU metric in Correlation Coeffition row, Sigma row represents the error rate (it should be less that 0,05) and N is number of samples taken into the experiment. The Table 7 provides results if Spearman's correlation for BLEU metric.

TABLE VI. SPEARMAN CORRELATION FOR EBLEU

|  | BLEU | NIST | TER | MET | RIB |
|---|---|---|---|---|---|
| Corr. Coefficient | 0.950 | 0.943 | -0.954 | 0.895 | 0.655 |
| Sigma (2-tailed) | 0.000 | 0.000 | 0.000 | 0.000 | 0.001 |
| N | 26 | 26 | 26 | 26 | 26 |

The Sigma of the Spearman correlation indicates the direction of association between *X* (the independent variable) and *Y* (the dependent variable). If *Y* tends to increase when *X* increases, the Spearman correlation coefficient is positive. If *Y* tends to decrease when *X* increases, the Spearman correlation coefficient is negative. A Spearman correlation of zero indicates that there is no tendency for *Y* to either increase or decrease when *X* increases. The Spearman correlation increases in magnitude as *X* and *Y* become closer to being perfect monotone functions of each other. When *X* and *Y* are perfectly monotonically related, the Spearman correlation coefficient becomes equal to 1.

For example -0.951 for TER and EBLEU shows strong negative correlation between these values. What is more, other results as well confirm strong and good correlations between measured metrics. Correlation between EBLEU *BLEU is equal to 0.947, for EBLEU *NIST result is 0.940, for EBLEU *TER is equal to -0.951 and for EBLEU *METEOR result is 0.891, which shows strong associations between these variables. The results for RIBES metric show rather moderate that very strong correlation.

TABLE VII. SPEARMAN CORRELATION FOR BLEU

|  | EBLEU | NIST* | TER* | MET* | RIB* |
|---|---|---|---|---|---|
| Corr. Coefficient | 0.950 | 0.915 | -0.945 | 0.897 | 0.655 |
| Sigma (2-tailed) | 0.000 | 0.000 | 0.000 | 0.000 | 0.001 |
| N | 26 | 26 | 26 | 26 | 26 |

In the other hand for BLEU metric we obtained following results, for BLEU*NIST 0.912, for BLEU*TER 0.939 and for BLEU*METEOR correlation coefficient is equal to 0.897 which shows strong association between variables as well but to as strong as EBLEU represents. Low correlation for RIBES occurs for each kind of translation.

## V. CONCLUSIONS AND FUTURE WORK

In our research we proved by measuring correlations that our variation of BLEU is trust worthier than normal BLEU. There are no deviations from the measurements from other metrics. Moreover our method of evaluation is more similar to human evaluation. We are assured with our experiments that our tool can provide better precision especially for Polish and other Slavic languages. As anticipated the correlation between our implementation and RIBES metric is not too strong. The focus of the RIBES metric is word order, which is free in Polish language. To be more precise it uses rank correlation coefficients based on word order to compare SMT and reference translations. As word order is free for polish – like languages having here rather weak correlation is a good sign.

The enhanced BLEU can deal with disparity of vocabularies between language pairs, and free word order that occurs in some none positional languages. We left in it an open gate for further adjustments in the final scores. The tool allows the changes in the proportions in which BLEU score is being altered with our enhancements. Thanks to that the tool can easily be adjusted to any language pairs or specific experimental needs.

## VI. ACKNOWLEDGEMENTS

This work is supported by the European Community from the European Social Fund within the Interkadra project UDA-POKL-04.01.01-00-014/10-00 and Eu-Bridge 7th FR EU project (grant agreement n°287658).

---

[1] http://onlinestatbook.com/2/describing_bivariate_data/pearson.html

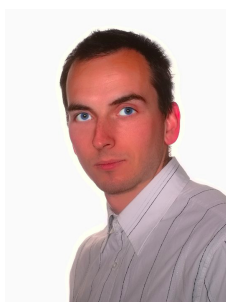

**Krzysztof Wolk** PHD student and Polish Japanese Institute of Information Technology.